# Online Sparse Streaming Feature Selection Using Adapted Classification

RuiYang Xu, Di Wu, *Member, IEEE*, Xin Luo, *Senior Member, IEEE*,

*Abstract*—Traditional feature selections need to know the feature space before learning, and online streaming feature selection (OSFS) is proposed to process streaming features on the fly. Existing methods divide features into relevance or irrelevance without missing data, and deleting irrelevant features may lead to information loss. Motivated by this, we focus on completing the streaming feature matrix and division of feature correlation and propose online sparse streaming feature selection based on adapted classification (OS$^2$FS-AC). This study uses Latent Factor Analysis (LFA) to pre-estimate missed data. Besides, we use the adaptive method to obtain the threshold, divide the features into strongly relevant, weakly relevant, and irrelevant features, and then divide weak relevance with more information. Experimental results on ten real-world data sets demonstrate that OS$^2$FS-AC performs better than state-of-the-art algorithms.

## I. Introduction

WITH the rapid development of information technology, the degree of interaction between data is gradually increasing [58], which increases data volume, fast growth rate, and diverse structure, resulting in the characteristics of multi-level, multi-granularity, multi-modality, and heterogeneity. Accordingly, artificial intelligence, communication, and storage technology are facing challenges [1]-[4]. For high-dimensional data, how to select appropriate features receive wide attention [5], [6].

Traditional feature selection methods focus on the filter [7] [8], wrapper [9], and embedded [10] [11], in which feature space is predefined or known. However, the feature space cannot predict in advance and may grow or even be infinite on the fly [12]-[14]. Therefore, many methods are proposed to deal with deals with streaming feature in an online manner. Two typical methods of streaming feature selection are OSFS [15] and SAOLA [16]. OSFS selects the weak relevance but non-redundancy and strong relevance features and contains two significant parts: online correlation analysis and online redundancy analysis. SAOLA considers the pairwise relationship of features calculated by mutual information in high-dimensional data. Therefore, many scholars proposed selected features methods applied to streaming features. However, existing streaming feature selection methods are applied in streaming features without miss data, and they may perform badly for sparse streaming features.

In practical applications, there are missing values in large-scale data. With the increase in data characteristics, some data cannot be collected, and the probability of missing data will increase. In bioinformatics, because of the limitation of cell sequencing technology, it is difficult to analyze the measured values of genes and cells when they reach a specific value [17]. ICU patient data is missing due to human factors and equipment failure [18]. In addition, data storage and conversion will also cause data loss [19]-[20]. Motivated by this, many scholars put forward the method of sparse matrix completion [21]-[26]. Wu *et al.* [27] proposed an online feature selection algorithm to complete the sparse matrix, named LOSSA. This method completes the sparse matrix, but there is always some error between the completed matrix and the actual value, which will reduce the accuracy of the feature selection. It is necessary to process the completed data to minimize the influence of errors on feature selection [28]-[31].

Existing approaches to online correlation analysis focus on removing irrelevant features and cannot classify features into weak relevance. Weak relevance is misclassified as an irrelevance and can not be recovered after deletion. Complete matrices with errors are more likely to be misclassified, and the accuracy of the LOSSA algorithm will be reduced. Inspired by the philosophy of three-way decision (3WD) [32], we can divide features into tri-partitions: selecting, discarding, and delaying. Weakly relevant features are divided into delayed decisions and waiting for more information.

Aiming to solve the above problems, we propose online sparse streaming feature selection via adapted classification (OS$^2$FS-AC), shown as Figure 1. We focus on selecting the weak relevance but non-redundancy and strong relevance features from sparse streaming data. Combined with the latent factor analysis (LFA) model, estimating the miss data. And then, based on the central idea of 'thinking in three,' the OS$^2$FS-AC automatically updates the values of $\beta$ and $\alpha$ to develop online correlation analysis. We summarize our main contributions can be summarized as follows:

a) The state-of-the-art competing online streaming feature selection algorithms usually use conventional methods such as zero-filling or average-filling when dealing with missing data. A significant gap between the completed data and the actual value affects the feature selection result. We use the LFA model to complete missing data to reduce the error.

b) In the correlation analysis, the correlation near the significance level may be misclassified, and the features will be discarded directly and cannot be used and selected again. To solve this problem, we require the three-way decision to put the weakly relevant features into the boundary region, waiting for sufficient conditions.

c) We handle the self-adaption strategy to update the correlation threshold automatically. The thresholds dynamically divided the features into strong relevance, weak relevance,

R. Xu is with School of Computer Science and Technology, Chongqing University of Posts and Telecommunications, Chongqing, 400065, China (e-mail: d220201035@stu.cqupt.edu.cn).

D. Wu and X. Luo are with the College of Computer and Information Science, Southwest University, Chongqing 400715, China (e-mail: wudi.cigit@gmail.com; luoxin@swu.edu.cn).

and irrelevance, improving correlation division accuracy.
d) To investigate the effectiveness between OS²FS-AC and six existing OSFS methods, we conduct experimental comparisons on 14 real-world microarray data sets. The experimental results indicate our new method performs better on predictive accuracy.

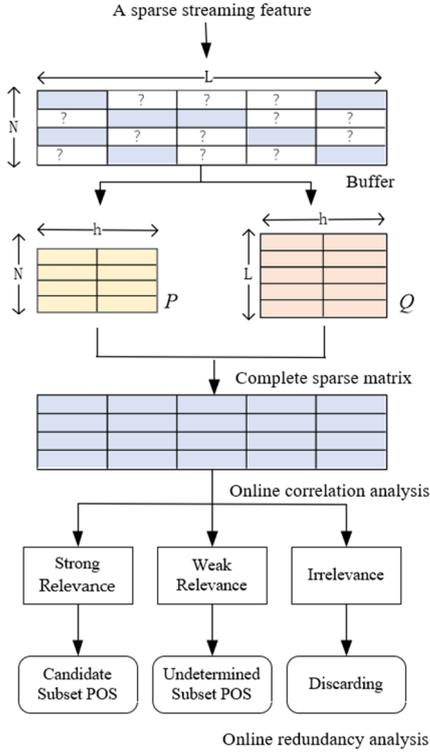

Fig.1. Flowchart of OS²FS-AC to achieve OS²FS.

## II. PRELIMINARIES

### A. Online Streaming Feature Selection

Given streaming features set $F = \{F_1, F_2, \cdots, F_{\psi_t}\}$, at timestamp $t (t \in \{1, 2, \cdots, \psi_t\})$, features $F_t = [f_{1,t}, f_{2,t}, \cdots, f_{N,t}]^T$ with $N$ instance. The point is to choose the weak relevance but non-redundancy and strong relevance features on the fly. We first define conditional independence as follows.

*Definition 1 (Conditional independence [15]):* Assuming two distinct features $F_t, F_l \in F$, $t \neq l$, $t, l \in \{1, 2, \cdots, \psi_t\}$, conditionally independent on a subset $S \subseteq F$, satisfy: $P(F_t | F_l, S) = P(F_t | S)$ or $P(F_t | F_l, S) = P(F_l | S)$.

In terms of conditional independence, we define strongly relevant, weakly relevant, and irrelevant features of class attribute $C = [c_1, c_2, \cdots, c_N]^T$ as follows:

*Definition 2 (Strongly Relevant, Weakly Relevant, and Irrelevant Feature, OSFS [15]):* At time stamp $t$, the inflow of a feature $F_t$.

a) A feature $F_t$ is strongly relevant to $C$, if $\forall S \subseteq F - \{F_t\}$
   s.t. $P(C | S, F_t) \neq P(C | S)$.

b) A feature $F_t$ is weakly relevant to $C$, if $\exists S \subseteq F - \{F_t\}$
   s.t. $P(C | S, F_t) \neq P(C | S)$.

c) A feature $F_t$ is irrelevant to $C$, if $\forall S \subseteq F - \{F_t\}$ s.t.
   $P(C | S, F_t) = P(C | S)$.

The existing online correlation analysis can only be divided into irrelevance and relevance (including strong and weak relevance). Still, it cannot distinguish between strong and weak relevance. Considering the correlation of features, we use the Markov blanket for redundancy analysis, which is given as follows:

*Definition 3 (Markov blanket [15]):* Supposed $X$ ($X \subset F$) is a Markov blanket for $C$, namely $M(C)$, satisfy:

$$\forall D \in F - X, P(C | X, D) = P(C | X). \quad (1)$$

We can remove redundant features of strong and weak relevance using the Markov blanket.

### B. Three-way decision (3WD)

In this paper, we adopt the three-way decision(3WD) to analyze the relevance of features, so taking a brief introduction to the 3WD. The core idea of 3WD is 'thinking in three,' that is, dividing a set into three regions.

*Definition 4 (Three-way decision [32]):* Given a finite non-empty set $W$, which can be divided into tri-partitions $\pi = \{POS, BND, NEG\}$, satisfies the following conditions:
a) $POS \cap BND = \emptyset, POS \cap NEG = \emptyset, BND \cap NEG = \emptyset$;
b) $POS \cup BND \cup NEG = W$.

These three regions can be totally ordered or partially ordered. It is possible to use one evaluation, two evaluation, or three evaluation models to construct the tri-partitions.

*Definition 5 (Three-way Classification [32]):* Assume that an evaluation $e: W \to (R, \prec)$ on $W$, where $e(x)$ denotes the evaluation value of $x$. Let a pair of thresholds $(\beta, \alpha)$ ($\alpha, \beta \in R$, with $\beta \prec \alpha$) divide three regions according to the following rules:

$$POS^{(\cdot, \alpha]}(e) = \{x \in W | e(x) \succeq \alpha\};$$

$$BND^{(\beta, \alpha)}(e) = \{x \in W | \beta < e(x) < \alpha\};$$

$$NEG^{[\beta, \cdot)}(e) = \{x \in W | e(x) \preceq \beta\}.$$

where $POS, BND, NEG$ represent positive, boundary, and negative region. 3WD applies different strategies depending on the characteristics of the three regions to improve decision-making effectiveness.

## III. PROPOSED ALGORITHM

### A. Problem of OS²FS-AC

Missing data of $F$ is sparse streaming features $F' = \{F'_1, F'_2, \cdots, F'_{\psi_t}\}$, the loss rate of $F'_t$ is $\zeta_t = 1 - |\Lambda_t| / N$, where the known values of $F'_t$ is $\Lambda_t$, $|\bullet|$ denotes the cardinality of the set, $\zeta$ represents the total loss rate of $F'$. At time stamp $t$, we get a new feature $F'_t$ with miss data. The challenge of streaming feature selection is to select a minimum subset $POS_{t-1} \cup \{F'_t\}$, so OS²FS-AC aims to minimize the decision cost of selected features.

For developing the sparse streaming features selection, three main difficulties need to be overcome: (1) how to use the LFA model to complete a sparse feature matrix with minimum error; (2) how to make an adaptive correlation analysis with the three-way decision; (3) how to analyze the redundancy of different related attributes.

In this article, we adopt Fisher's Z-test to calculate correlation between continuous features for completed streaming feature matrix $\widehat{U}$. Given a feature subset $S$, the independence between features $F_t$ and $Y$ can be computed as follows:

$$P(F_t, Y|S) = \frac{-\left(\left(\Sigma_{F_tYS}\right)^{-1}\right)_{F_tY}}{\sqrt{\left(\left(\Sigma_{F_tYS}\right)^{-1}\right)_{F_tF_t}\left(\left(\Sigma_{F_tYS}\right)^{-1}\right)_{YY}}} \quad (2)$$

Under the null hypothesis of the conditional independence between $C$ and $F_t$, with $\rho-value$ returned by Fisher's Z-test to measure conditional independence and a significance level, $\mu$ judges the correlation of a feature, dividing features $F_t$ into relevance or irrelevance. If $\rho \leq \mu$, feature $F_t$ is divided into the relevant feature set $\mathbb{C}$; on the contrary, if $\rho > \mu$, then feature $F_t$ is divided into irrelevant feature set $\neg\mathbb{C}$.

*B. OS²FS-AC*

To make OS²FS-AC as flexible as possible to improve existing OSFS algorithms to handle OS²FS, we design its processing flow, as shown in Fig. 1. OS²FS-AC has three phases. Phase I pre-processes sparse streaming features to complete their missing data. Phase II performs online correlation analysis. Next, Phase III performs online redundancy analysis.

*1) Phase I Complete streaming feature matrix*

With the continuous inflow of streaming features, a buffer is a temporarily stored feature. After the feature dimension reaches a specific value, the missing data in the buffer will be completed. From time stamp $t$ to $t+L-1$, supposing sparse streaming feature flow in $U^{N \times L}$ buffer, denoted by $U = \{F'_t, F'_{t+1}, \cdots, F'_{t+L+1}\}$, where $N$ represents the sample size, and the column number is $L$. Use LFA [33]-[39] model to supplement the miss data of $U^{N \times L}$ buffer relying on the known data.

*Definition 6 (LFA [40], [41]).* The known data in sparse matrix $U^{W \times Z}$ denote as $\Lambda_U$, and the rank-h approximation $\widehat{U} = PQ^T$ for $U$ is estimated depending on the $\Lambda_U$; that is, the predicted missing value of two latent factor matrices $P^{W \times h}$ and $Q^{Z \times h}$ is obtained by training, and $h$ is the latent factor dimension; $h \ll \min\{|W|, |Z|\}$.

LFA model tries to build a low-rank approximation $\widehat{U} = \{\widehat{F}'_t, \widehat{F}'_{t+1}, \cdots, \widehat{F}'_{t+L+1}\}$ to a sparse streaming feature matrix $U$. Commonly, to obtain matrix $P$ and $Q$ from $\Lambda_U$, the minimum loss function is constructed by Euclidean distance between $U$ and $\widehat{U}$; we have

$$\varepsilon(P,Q) = \frac{1}{2} \sum_{f'_{n,j} \in \Lambda_U} \left( f'_{n,j} - \sum_{k=1}^{h} p_{n,k}q_{j,k} \right)^2 + L, \quad (3)$$

where $f'_{n,j} \in \Lambda_U$ is the known value, $p_{n,k}$ is the $n$-th row and $k$-th column of $P$, and $q_{j,k}$ is the $j$-th row and $k$-th column of $Q$, $j \in \{t, t+1, \cdots, t+L-1\}$, $\forall n \in \{1,2,\cdots,N\}$, $\forall k \in \{1,2,\cdots,h\}$. $L$ is a regularization scheme to avoid overfitting. The $L_2$ scheme is used to improve the generalization ability [42]-[46], and the calculation formula is formulated by:

$$L = \frac{\lambda}{2}(\|P\|_F^2 + \|Q\|_F^2), \quad (4)$$

where $\|\cdot\|_F$ computes the Frobenius norm, $\odot$ is the Hadmard product, and $\lambda$ denotes the regularization parameter.

Based on formula (3) and (4), predict the miss data acording to the known value is given by:

$$\varepsilon = \sum_{f'_{n,j} \in \Lambda_U} \left( \frac{1}{2}\left(f'_{n,j} - \sum_{k=1}^{h} p_{n,k}q_{j,k}\right)^2 + \frac{\lambda}{2}\left(\sum_{k=1}^{h} p_{n,k}^2 + \sum_{k=1}^{h} q_{j,k}^2\right)\right) \quad (5)$$

The loss function of the $n$-th element $f'_{n,j}$ of $F'_j$:

$$\varepsilon_{n,j} = \frac{1}{2}\left(f'_{n,j} - \sum_{k=1}^{h} p_{n,k}q_{j,k}\right)^2 + \frac{\lambda}{2}\left(\sum_{k=1}^{h} p_{n,k}^2 + \sum_{k=1}^{h} q_{j,k}^2\right). \quad (6)$$

We use stochastic gradient descent (SGD) to solve the loss function [47]-[50], calculate the gradient of the loss function to the sum of parameters, and update it in the negative gradient direction of parameters $p_{n,k}$ and $q_{j,k}$. The formula of $p_{n,k}$ and $q_{j,k}$:

$$\begin{cases} p_{n,k} \leftarrow p_{n,k} - \eta \frac{\partial \varepsilon_{n,j}}{\partial u_{n,k}} \\ q_{j,k} \leftarrow q_{j,k} - \eta \frac{\partial \varepsilon_{n,j}}{\partial u_{n,k}} \end{cases}. \quad (7)$$

In terms of formulas (6) and (7), the partial derivative of the loss function is obtained:

$$\begin{aligned} p_{n,k} &\leftarrow p_{n,k} + \eta q_{j,k}\left(f'_{n,j} - \sum_{k=1}^{h} p_{n,k}q_{j,k}\right) - \lambda\eta p_{n,k} \\ q_{j,k} &\leftarrow q_{j,k} + \eta p_{n,k}\left(f'_{n,j} - \sum_{k=1}^{h} p_{n,k}q_{j,k}\right) - \lambda\eta q_{j,k}. \end{aligned} \quad (8)$$

where $\eta$ is the learning rate, and two late factor matrices $P$ and $Q$ are trained with the estimation of the minimum errors on the known value, then $\widehat{U} = PQ^T$. To make the formula (8) more concise, the error between the predicted value and the actual value is recorded as $err_{n,j} = f'_{n,j} - \sum_{k=1}^{h} p_{m,k}q_{j,k}$, that is,

$$\begin{aligned} p_{n,k} &\leftarrow p_{n,k} + \eta q_{j,k}err_{n,j} - \lambda\eta p_{n,k} \\ q_{j,k} &\leftarrow q_{j,k} + \eta p_{n,k}err_{n,j} - \lambda\eta q_{j,k}. \end{aligned} \quad (9)$$

*2) Phase II Online correlation analysis*

Most of the existing correlation analysis methods use two-way decision(2WD); $a_P$ and $a_E$ represent two actions of acceptance and rejection, which needs to analyze the feature relevance immediately. However, the fluctuation of feature relevance near the threshold $\mu$ will lead to misclassification. Three-way decision moves the misclassified samples in 2WD to the boundary region as much as possible and waits for the information to further process the features of the boundary region. Compared with relevant feature and irrelevant feature, we use a new option to effectively improve the accuracy of feature selection. Action $a_B$ is a delayed decision, so the action set $A = (a_P, a_B, a_E)$ re-divides feature into strong relevance, weak relevance, or irrelevance. The action set $A=(a_P,a_B,a_E)$ describes the problem of feature relevance partition, and the relevance feature set $[\mathbb{C},\neg\mathbb{C}]$ is divided into three partitions. Different divisions will cause the corresponding cost in this process, and $r_{PP},r_{BP},r_{EP}$ denotes the costs incurred for applying strategies $a_P,a_B,a_E$, respectively, when a feature belongs to the relevant feature set $\mathbb{C}$. Analogously, $r_{PE},r_{BE},r_{EE}$ denotes the costs incurred for applying strategies $a_P,a_B,a_E$, respectively, when a feature belongs to the irrelevant feature set $\neg\mathbb{C}$. We assume misclassification costs are more than the correct classification, that is, $r_{PP}\leq r_{BP}\leq r_{EP}$ and $r_{EE}\leq r_{BE}\leq r_{PE}$ [51]. The cost matrix is shown in the Table I.

TABLE I COST MATRIX.

| Action | Cost Function | |
|---|---|---|
| | $\mathbb{C}$ | $\neg\mathbb{C}$ |
| $a_P$ | $r_{PP}$ | $r_{PE}$ |
| $a_B$ | $r_{BP}$ | $r_{BE}$ |
| $a_E$ | $r_{EP}$ | $r_{EE}$ |

The relevance between feature $F_t$ and class attribute $C$ is $Dep(\mathbb{C},F_t)$, and the irrelevance between feature $F_t$ and class attribute $C$ is $Dep(\neg\mathbb{C},F_t)$. Accordingly, the expected cost associated with applying strategies $a_P,a_B,a_E$ be computed by:

$$R(a_P\,|\,[F_t])=r_{PP}Dep(\mathbb{C},F_t)+r_{PE}Dep(\neg\mathbb{C},F_t),$$
$$R(a_B\,|\,[F_t])=r_{BP}Dep(\mathbb{C},F_t)+r_{BE}Dep(\neg\mathbb{C},F_t),$$
$$R(a_E\,|\,[F_t])=r_{EP}Dep(\mathbb{C},F_t)+r_{EE}Dep(\neg\mathbb{C},F_t).$$

where $R(a_\cdot\,|\,[F_t])(\cdot\in\{P,B,E\})$ denote the cost of feature $F_t$ when taking action $a_\cdot(\cdot\in\{P,B,E\})$.

According to the Bayesian decision rule, induced by the minimum-cost decision rules, the decision rules can be given by:

P): If $R(a_P\,|[F_t])\leq R(a_B\,|[F_t])$ and $R(a_P\,|[F_t])\leq R(a_E\,|[F_t])$, decide $F_t\in POS_t$, then feature $F_t$ is divided into strong relevance;

B): If $R(a_B\,|[F_t])\leq R(a_P\,|[F_t])$ and $R(a_B\,|[F_t])\leq R(a_E\,|[F_t])$, decide $F_t\in BND_t$, then feature $F_t$ is divided into weak relevance;

E): If $R(a_E\,|[F_t])\leq R(a_P\,|[F_t])$ and $R(a_B\,|[F_t])\leq R(a_E\,|[F_t])$, decide $F_t\in NEG_t$, then feature $F_t$ is divided into irrelevance.

In addition, the decision rules procedure suggests the following simplified feature relevance classification rules:
a) If $Dep(\mathbb{C},F_t)\geq\alpha$, decide $F_t\in POS_t$, then feature $F_t$ is divided into strong relevance;
b) If $\beta<Dep(\mathbb{C},F_t)<\alpha$, decide $F_t\in BND_t$, then feature $F_t$ is divided into weak relevance;
c) If $Dep(\mathbb{C},F_t)\leq\beta$, decide $F_t\in NEG_t$, then feature $F_t$ is divided into irrelevance.

And then, consider the cost of the three-way decision of feature relevance, so we define the cardinal number of the above six conditions as follows:

TABLE II CARDINAL NUMBER.

| Action | Cost Function | |
|---|---|---|
| | $\mathbb{C}$ | $\neg\mathbb{C}$ |
| $a_P$ | $m_{PP}^t$ | $m_{PE}^t$ |
| $a_B$ | $m_{BP}^t$ | $m_{BE}^t$ |
| $a_E$ | $m_{EP}^t$ | $m_{EE}^t$ |

where action $a_P,a_B,a_E$ divide the features in set $\mathbb{C}$ into the total number of $m_{PP}^t,m_{BP}^t,m_{EP}^t$, respectively. Similarly, action $a_P,a_B,a_E$ divide the features in set $\neg\mathbb{C}$ into the total number of $m_{PE}^t,m_{BE}^t,m_{EE}^t$, respectively.

Obviously, at time stamp $t$, the decision cost of three-way correlation analysis can be computed as:

$$COST_t=r_{EP}m_{EP}^t+r_{PE}m_{PE}^t+r_{BP}m_{BP}^t+r_{BE}m_{BE}^t+r_{PP}m_{PP}^t+r_{EE}m_{EE}^t. \quad (10)$$

For convenience, extending from the assumptions of [51], we define the cost for the right classification as zero, that is, $r_{PP}=r_{NN}=0$. Based on formula (10), predict the miss data according to the known value, and the cost function is given by:

$$COST_t=r_{EP}m_{EP}^t+r_{PE}m_{PE}^t+r_{BP}m_{BP}^t+r_{BE}m_{BE}^t, \quad (11)$$

where $COST_t^{mis}=r_{EP}m_{EP}^t+r_{PE}m_{PE}^t$ denote the misclassification cost, $COST_t^{del}=r_{BP}m_{BP}^t+r_{BE}m_{BE}^t$ denote the delayed classification cost.

Three-way correlation analysis needs to be divided in terms of the threshold value, and the initial threshold value $\alpha,\beta$ is calculated according to the literature [52]:

$$\alpha=\frac{(r_{PE}-r_{BE})}{(r_{PE}-r_{BE})+(r_{BP}-r_{PP})},$$
$$\beta=\frac{(r_{BE}-r_{EE})}{(r_{BE}-r_{EE})+(r_{EP}-r_{BP})}. \quad (12)$$

To ensure the accuracy of feature relevance division, it is essential to set appropriate thresholds $(\alpha_t,\beta_t)$ at every moment. Finding an optimal threshold for feature $F_t$ and meeting the threshold with the lowest risk cost in decision-making, the problem of solving threshold parameters becomes an optimization problem, namely:

$$\underset{(\alpha_t,\beta_t)}{\arg\min}(COST_t). \quad (13)$$

We use a simulated annealing algorithm to solve the optimal threshold. Calculate the initial threshold according to formula (12), and automatically update the threshold $(\alpha_t,\beta_t)$ in each iteration. If the threshold update reduces or keeps the decision cost $COST_t$ unchanged, the threshold is updated; otherwise, the threshold update is accepted with a certain probability, and the iterative process is repeated until the minimum decision cost is reached. The algorithm is given in Table III. In this algorithm, The temperature change is $nowT=nowT*delta$, where $delta=0.95$ regulates the temperature.

TABLE III THREE-WAY CORRELATION THRESHOLD ALGORITHM BASED ON SIMULATED ANNEALING.

```
1    initialize r_{PP}, r_{BP}, r_{EP}, r_{PE}, r_{BE}, r_{EE}; initT, minT, delta
2    get a streaming feature F_t at time stamp t
3    nowT = initT
4    initialize α_t, β_t according to (12)
5    initialize COST_t according to (11)
6    α_t^{new} = α_t, β_t^{new} = β_t
7    while nowT < minT
8        update α_t^{new}, β_t^{new}
9        if 0 ≤ β_t^{new} < α_t^{new} ≤ 1
10           calculate nowCOST_t according to ()
11       else
12           update α_t^{new}, β_t^{new} until 0 ≤ β_t^{new} < α_t^{new} ≤ 1
13       end if
14       if COST_t − nowCOST_t > 0
15           COST_t = nowCOST_t
16           α_t = α_t^{new}, β_t = β_t^{new}
17       else
18           if random.rand < exp(−res/(k*nowT))
19               COST_t = nowCOST_t
20               α_t = α_t^{new}, β_t = β_t^{new}
21           end if
22       end if
23       nowT = nowT * delta
24   end while
```

*3) Phase III Online redundancy analysis*

After online correlation analysis, we need to carry on a redundant analysis depending on the correlation of features, relying on $\rho-value$ returned by Fisher's Z-test to calculate conditional dependence $Ind(C, F_t | X_F)$. At time stamp $t$, let $M(C)_t$ represents a Markov blanket of $C$ at time stamp $t$, with correlation analysis, a feature $F_t$ is relevant to $C$, if $\exists X_F \in M(C)_t$, s.t. $P(C|F_t, X_F) = P(C|F)$, which concludes that $F_t$ is a redundant feature. Based on the Markov blanket, we propose proposition 1 to judge whether attributes are redundant.

**Proposition 1.** For strongly relevant feature $F_t$, add them to $POS_t$ and conduct a redundancy analysis between $F_t$ and features in $POS_t$; if

$$\exists X_F \in POS_t \text{ s.t. } P(C|F_t, X_F) = P(C|X_F), \quad (14)$$

then the feature $F_t$ is redundant can be discarded. For weakly relevant feature $F_t$, when $POS_{t-1}$ is not an empty set, if

$$\forall X_F \in POS_{t-1} \text{ s.t. } P(C|F_t, X_F) \neq P(C|X_F), \quad (15)$$

then the feature $F_t$ is non-redundant and added to $POS_t$. When $POS_{t-1}$ is an empty set, the weakly relevant feature $F_t$ is put into $BND_t$ until $POS_{t-1}$ is not an empty set, then the feature $F_i \in POS_{t-1}$ and $F_j \in BND_t$ are analyzed for redundancy.

Non-redundant features are added into $POS_t$, which may lead to the redundancy of the original features in $POS_t$, so we propose proposition 2 to check the redundancy of the other features in $POS_t$.

**Proposition 2.** At time point, if a feature $F_t$ flow in, $M(F_t) \notin M(C)_t$,

$$\forall X_F \in M(C) \cup F_t, \exists \vartheta \subseteq M(C) \cup F_t - \{X_F\} \text{ s.t.}$$
$$P(C|X_F, \vartheta) = P(C|\vartheta) \quad (16)$$

then $\{X_F\}$ is redundant and should be deleted from $POS_t$.

## IV. EXPERIMENTS AND RESULTS

### A. General Settings

**Datasets.** In this section, we use 10 real-world data sets form DNA microarray data, NIPS 2003 data, public microarray data and studies in [53], [54] as show in Table IV.

TABLE IV THE DETAILS OF SELECTED DATASETS.

| Mark | Dataset | #(Features) | #(Instances) | #(Class) |
|---|---|---|---|---|
| D1 | USPS | 1500 | 242 | 2 |
| D2 | Colon | 2001 | 62 | 2 |
| D3 | SRBCT | 2309 | 83 | 4 |
| D4 | Lung | 3313 | 83 | 5 |
| D5 | Prostate | 6033 | 102 | 2 |
| D6 | Leukemia | 7071 | 72 | 2 |
| D7 | Lungcancer | 12534 | 181 | 2 |
| D8 | SMK-CAN-187 | 19993 | 187 | 2 |
| D9 | Madelon | 501 | 2600 | 6 |
| D10 | HAPT | 561 | 10929 | 12 |

**Baselines.** We choose five state-of-the-art streaming feature selections to validate our algorithm performance, including Fast-OSFS [15], SAOLA [16], SFS-FI [55], OSSFS-DD [56], and LOSSA [27]. Besides, we adopt three basic classifiers, SVM, KNN, and random forest, to evaluate the validity of feature selection. Table V summarizes the parameters set to the values of these classifiers and algorithms. These algorithms are conducted on MATLAB [57] We perform 5-fold cross-validation in our experiments, where 4/5 data is the training set, and 1/5 data is the test set. Repeating each data ten times and reporting the predictive accuracy, mean number of selected features, and the running time. All experiments are implemented on a personal computer (Intel i7 2.40-GHz CPU, RAM 16GB).

TABLE V ALL THE PARAMETERS USED IN THE EXPERIMENTS.

| Mark | Algorithm | Parameter |
|---|---|---|
| M1 | OS²FS-AC | Z test, Alpha is 0.05. |
| M2 | Fast-OSFS | Z test, Alpha is 0.05. |
| M3 | SAOLA | Z test, Alpha is 0.05. |
| M4 | SFS-FI | Z test, Alpha is 0.05, $\gamma = 0.05$. |
| M5 | OSSFS-DD | $k_1 = 2$, $k_2 = 3$, $N = 2.35\% * n$ ($n$ is number of attributes) |
| M6 | LOSSA | Z test, Alpha is 0.05, $\lambda = 0.01$, $\eta = 0.01$, $B_S = 10$ |
| / | KNN | The neighbors are 3. |

**Experimental Designs.** We compared the OS²FS-AC algorithm with the algorithms above on sparse streaming features with 10% missing data. The default setting is: $\lambda = 0.01$, $\eta = 0.01$, $L = 5$, $h = 10$, and the cost matrix adopted by our algorithm is $r_{PP} = 0$, $r_{BP} = 1$, $r_{EP} = 10$, $r_{PE} = 10$, $r_{BE} = 1$, $r_{EE} = 0$. To further check the performance of the different algorithms, we conduct the Friedman test at a 95% significance level under the null hypothesis.

## B. OS²FS-AC vs. Online Streaming Feature Selection Methods

From Table VI, we can see OS²FS-AC gets lower average ranks and higher average accuracy than other algorithms in the cases of KNN, SVM, and random forest. And we have the following observations:

a) OS²FS-AC Versus Fast-OSFS: For example, on data sets HAPT, SFS-FI selects 102 features while getting lower accuracy than OS²FS-AC. The main reason is that Fast-OSFS incompletes sparse matrix and employs two-way classification on correlation analysis that causes misclassification, leading to inferior accuracy performance.

b) OS²FS-AC Versus SAOLA: SAOLA only considers the feature relationships between two features, which causes some vital information loss. However, with the full use of the LFA model and three-way decision, OS²FS-AC can always select the critical informative features on the fly while having better predictive accuracy.

c) OS²FS-AC Versus SFS-FI: SFS-FI can not select any features for 10% missing data sets and only select the first features on some data sets, such as Lungcancer and SMK-CAN-187, causing the loss of critical information. Therefore, SFS-FI cannot handle sparse streaming features well.

d) OS²FS-AC Versus OSSFS-DD: OSSFS-DD only has a predictive accuracy of around 0.9 on Leukemia and 0.2 on Madelon. The unsteady average accuracy is that the OSSFS-DD algorithm also adopts three-way classification in correlation analysis, but the threshold is only dynamically updated by calculation. Unlike the automatic updating of the threshold in this paper, the updating method of OS2FS-AC is universal. This demonstrates that OS2FS-AC is more stable than OSSFS-DD.

e) OS²FS-AC Versus LOSSA: Though the sparse matrix is completed, LOSSA uses the two-way classification for correlation analysis that cannot select features well. Meanwhile, it does not deal with redundancy between a new feature and selected features, which causes lower average accuracy.

TABLE VI USING THE SELECTED FEATURES (RECORDED IN TABLE V) TO TRAIN A CLASSIFIER FIRST AND THE TESTING ITS ACCURACY (%), $\alpha$=0.1.

| Models/Datasets | | D1 | D2 | D3 | D4 | D5 | D6 | D7 | D8 | D9 | D10 | Average | ^Rank |
|---|---|---|---|---|---|---|---|---|---|---|---|---|---|
| KNN | M1 | **88.46**$_{\pm 0.43}$ | **84.97**$_{\pm 0.20}$ | **85.56**$_{\pm 3.24}$ | **88.23**$_{\pm 1.40}$ | **94.54**$_{\pm 1.71}$ | **98.90**$_{\pm 0.77}$ | **98.89**$_{\pm 0.31}$ | 70.14$_{\pm 2.78}$ | **58.10**$_{\pm 0.50}$ | **77.56**$_{\pm 0.36}$ | **84.54**$_{\pm 1.17}$ | **1.20** |
| | M2 | 86.80$_{\pm 0.27}$ | 77.94$_{\pm 2.46}$ | 81.93$_{\pm 0.70}$ | 81.13$_{\pm 0.85}$ | 92.04$_{\pm 1.68}$ | 98.33$_{\pm 0.05}$ | 98.07$_{\pm 0.33}$ | 79.27$_{\pm 1.73}$ | 56.40$_{\pm 0.14}$ | 71.38$_{\pm 0.08}$ | 82.33$_{\pm 0.83}$ | 2.80 |
| | M3 | 82.61$_{\pm 0.49}$ | 80.96$_{\pm 2.89}$ | 82.88$_{\pm 2.00}$ | 82.06$_{\pm 0.79}$ | 91.88$_{\pm 0.55}$ | 98.74$_{\pm 0.00}$ | 98.45$_{\pm 0.00}$ | **86.37**$_{\pm 1.62}$ | 54.47$_{\pm 0.45}$ | 59.52$_{\pm 0.34}$ | 81.80$_{\pm 0.91}$ | 2.60 |
| | M4 | 74.41$_{\pm 0.59}$ | 64.51$_{\pm 0.68}$ | 57.49$_{\pm 0.79}$ | 63.70$_{\pm 1.03}$ | 73.51$_{\pm 3.46}$ | 34.65$_{\pm 0.20}$ | 87.69$_{\pm 1.80}$ | 61.78$_{\pm 1.07}$ | 49.76$_{\pm 0.22}$ | 50.10$_{\pm 0.37}$ | 61.76$_{\pm 1.02}$ | 5.40 |
| | M5 | 74.01$_{\pm 0.60}$ | 62.46$_{\pm 0.52}$ | 53.43$_{\pm 1.32}$ | 75.87$_{\pm 2.00}$ | 55.10$_{\pm 1.76}$ | 97.36$_{\pm 0.77}$ | 97.19$_{\pm 0.97}$ | 66.75$_{\pm 2.35}$ | 49.69$_{\pm 0.42}$ | 25.60$_{\pm 0.30}$ | 65.75$_{\pm 1.10}$ | 5.50 |
| | M6 | 85.47$_{\pm 0.56}$ | 74.32$_{\pm 0.74}$ | 82.66$_{\pm 1.87}$ | 78.34$_{\pm 0.49}$ | 90.77$_{\pm 2.04}$ | 96.93$_{\pm 0.00}$ | 98.12$_{\pm 0.01}$ | 68.99$_{\pm 1.36}$ | 54.77$_{\pm 0.17}$ | 72.40$_{\pm 0.18}$ | 80.28$_{\pm 0.74}$ | 3.50 |

## V. CONCLUSIONS

In this article, we study the problem of sparse streaming feature selection and propose a new online sparse streaming feature selection method via adapted classification (OS2FS-AC). The LFA model supplements the sparse matrix. At the same time, the attributes are divided into strong relevance, weak relevance, and irrelevant, with the help of three-way decisions, improving the accuracy of feature selection. In our further work, we will further study the method of completing a sparse matrix and combine various methods to carry out online sparse streaming feature selection.

## REFERENCES

[1] X. Luo, Hao Wu, Zhi Wang, Jianjun Wang, and Deyu Meng, "A Novel Approach to Large-Scale Dynamically Weighted Directed Network Representation," *IEEE Transactions on Pattern Analysis and Machine Intelligence,* 2022, 44(12): 9756-9773.

[2] S. Li, M. Zhou, X. Luo, and Z. -H. You, "Distributed Winner-Take-All in Dynamic Networks," *IEEE Transactions on Automatic Control,* vol. 62, no. 2, pp. 577-589, 2017.

[3] S. Li, M. Zhou and X. Luo, "Modified Primal-Dual Neural Networks for Motion Control of Redundant Manipulators With Dynamic Rejection of Harmonic Noises," *IEEE Transactions on Neural Networks and Learning Systems,* vol. 29, no. 10, pp. 4791-4801, 2018.

[4] S. Li, Z. -H. You, H. Guo, X. Luo, and Z. -Q. Zhao, "Inverse-Free Extreme Learning Machine With Optimal Information Updating," *IEEE Transactions on Cybernetics,* vol. 46, no. 5, pp. 1229-1241, 2016.

[5] G. Chandrashekar, and F. Sahin, "A survey on feature selection methods," *Computers and Electrical Engineering,* vol. 40, no. 1, pp. 16-28, 2014.

[6] S. Alelyani, J. Tang, and H. Liu, "Feature selection for clustering: a review," *Data Clustering*, pp. 29-60: Chapman and Hall/CRC, 2018.

[7] P. P. Kundu and S. Mitra, "Feature selection through message passing," *IEEE Trans. Cybern.,* vol. 47, no. 12, pp. 4356–4366, 2017.

[8] Y. Yang, D. Chen, H. Wang, and X. Wang, "Incremental perspective for feature selection based on fuzzy rough sets," *IEEE Trans. Fuzzy Syst.,* vol. 26, no. 3, pp. 1257–1273, Jun. 2018.

[9] X. Xue, M. Yao, and Z. Wu, "A novel ensemble-based wrapper method for feature selection using extreme learning machine and genetic algorithm," *Knowl. Inf. Syst.,* vol. 57, no. 2, pp. 389–412, 2018.

[10] H. Liu and H. Motoda, Computational Methods of Feature Selection. London, U.K.: Chapman & Hall, 2007.

[11] B. Xue, M. Zhang, W. N. Browne, and X. Yao, "A survey on evolutionary computation approaches to feature selection," *IEEE Trans. Evol. Comput.,* vol. 20, no. 4, pp. 606–626, 2016.

[12] J. Ni, H. Fei, W. Fan, and X. Zhang, "Automated medical diagnosis by ranking clusters across the symptom-disease network," *In Proceedings of the 2017 IEEE International Conference on Data Mining*, 2017, pp. 1009-1014.

[13] G. Ditzler, J. LaBarck, J. Ritchie, G. Rosen, and R. Polikar, "Extensions to online feature selection using bagging and boosting," *IEEE Trans. Neural Netw. Learn. Syst.,* vol. 29, no. 9, pp. 4504–4509, 2018.

[14] Y. Shen, C. Wu, C. Liu, Y. Wu, and N. Xiong, "Oriented feature selection SVM applied to cancer prediction in precision medicine," *IEEE Access,* vol. 6, pp. 48510-48521, 2018.

[15] X. Wu, K. Yu, W. Ding, H. Wang, and X. Zhu, "Online feature selection with streaming features," *IEEE Transactions on Pattern Analysis and Machine Intelligence,* vol. 35, no. 5, pp. 1178-1192, 2013.

[16] K. Yu, X. Wu, W. Ding, and J. Pei, "Scalable and accurate online feature selection for big data," *ACM Transactions on Knowledge Discovery from Data,* vol. 11, no. 2, pp. 16, 2016.

[17] M. B. Badsha, R. Li, B. X. Liu, et al. "Imputation of single-cell gene expression with an autoencoder neural network," *Quantitative Biology,* vol. 8, no. 1, pp. 78-94, 2020.

[18] A. Idri, H, Benhar, J. L. Fernández-Alemán, et al. "A systematic map of medical data preprocessing in knowledge discovery," *Computer Methods and Programs Biomedicine,* vol. 162, pp. 69-85, 2018.

[19] N. Zeng, P. Wu, Z. Wang, H. Li, W. Liu, X. Liu, "A small-sized object detection oriented multi-scale feature fusion approach with application to defect detection," *IEEE Transactions on Instrumentation and Measurement,* vol. 71, article no. 3507014, 2022.

[20] H. Li, P. Wu, N. Zeng, Y. Liu, Fuad E. Alsaadi, "A Survey on Parameter Identification, State Estimation and Data Analytics for Lateral Flow Immunoassay: from Systems Science Perspective," *International Journal of Systems Science,* DOI:10.1080/00207721.2022.2083262.

[21] X. Luo, Y. Yuan, S. L. Chen, N. Y. Zeng, and Z. D. Wang, "Position-Transitional Particle Swarm Optimization-Incorporated Latent Factor Analysis," *IEEE Transactions on Knowledge and Data Engineering,* vol. 34, no. 8, pp. 3958-3970, 2022.

[22] X. Luo, H. Wu, and Z. C. Li, "NeuLFT: A Novel Approach to Nonlinear Canonical Polyadic Decomposition on High-Dimensional Incomplete Tensors," *IEEE Transactions on Knowledge and Data Engineering,* DOI: 10.1109/TKDE.2022.3176466.

[23] X. Luo, Yue Zhou, Zhigang Liu, and MengChu Zhou, "Fast and Accurate Non-negative Latent Factor Analysis on High-dimensional and Sparse Matrices in Recommender Systems," IEEE Transactions on Knowledge and Data Engineering, DOI: 10.1109/TKDE.2021.3125252.

[24] D. Wu, X. Luo, M. S. Shang, Y. He, G. Y. Wang, and X. D. Wu, "A Data-Characteristic-Aware Latent Factor Model for Web Services QoS Prediction," *IEEE Transactions on Knowledge and Data Engineering,* vol. 34, no. 6, pp. 2525-2538, 2022.

[25] X. Luo, Y. R. Zhong, Z. D. Wang, and M. Z. Li, "An Alternating-direction-method of Multipliers-Incorporated Approach to Symmetric Non-negative Latent Factor Analysis," *IEEE Transactions on Neural Networks and Learning Systems,* DOI: 10.1109/TNNLS.2021.3125774.

[26] X. Luo, Y. Zhou, Z. G. Liu, L. Hu, and M. C. Zhou, "Generalized Nesterov's Acceleration-incorporated, Non-negative and Adaptive Latent Factor Analysis," *IEEE Transactions on Services Computing*, vol. 15, no.5, pp. 2809-2823, 2022.

[27] D. Wu, Y. He, X. Luo, and M. Zhou, "A Latent Factor Analysis-Based Approach to Online Sparse Streaming Feature Selection," *IEEE Transactions on Systems, Man, and Cybernetics: Systems,* vol. 52, no. 11, pp. 6744-6758, 2022.

[28] X. Luo, M. C. Zhou, Z. D. Wang, Y. N. Xia, and Q. S. Zhu, "An Effective Scheme for QoS Estimation via Alternating Direction Method-Based Matrix Factorization," *IEEE Transactions on Services Computing,* vol. 12, no. 4, pp. 503-518, 2019.

[29] F. H. Bi, X. Luo, B. Shen, H. L. Dong, and Z. D. Wang, "Proximal Alternating-Direction-Method-of-Multipliers-Incorporated Nonnegative Latent Factor Analysis," *IEEE/CAA Journal of Automatica Sinica,* DOI: 10.1109/JAS.2023.123474.

[30] D. Wu, M. S. Shang, X. Luo, and Z. D. Wang, "An L1-and-L2-norm-oriented Latent Factor Model for Recommender Systems," *IEEE Transactions on Neural Networks and Learning Systems,* vol. 33, no. 10, pp. 5775-5788, 2022.

[31] D. Wu, X. Luo, Y. He and M. C. Zhou, "A Prediction-sampling-based Multilayer-structured Latent Factor Model for Accurate Representation to High-dimensional and Sparse Data," *IEEE Transactions on Neural Networks and Learning Systems,* DOI: 10.1109/TNNLS.2022.3200009

[32] Y. Y. Yao, "The geometry of three-way decision," *Applied Intelligence,* pp. 1–28, 2021.

[33] F. H. Bi, T. T. He, Y. T. Xie, and Xin Luo, "Two-Stream Graph Convolutional Network-Incorporated Latent Feature Analysis," *IEEE Transactions on Services Computing,* DOI: 10.1109/TSC.2023.3241659.

[34] J. Chen, X. Luo, and M. C. Zhou, "Hierarchical Particle Swarm Optimization-incorporated Latent Factor Analysis for Large-Scale Incomplete Matrices," *IEEE Transactions on Big Data,* vol. 8, no. 6, pp. 1524-1536, 2022.

[35] Y. Yuan, X. Luo, M. S. Shang, and Z. D. Wang, "A Kalman-Filter-Incorporated Latent Factor Analysis Model for Temporally Dynamic Sparse Data," *IEEE Transactions on Cybernetics,* DOI: 10.1109/TCYB.2022.3185117.

[36] X. Luo, M. C. Zhou, S. Li, Z. H. You, Y. N. Xia, and Q. S. Zhu, "A Nonnegative Latent Factor Model for Large-Scale Sparse Matrices in Recommender Systems via Alternating Direction Method," IEEE Transactions on Neural Networks and Learning Systems, vol. 27, no.3, pp. 524-537, 2016.

[37] Z. G. Liu, X. Luo, and M. C. Zhou, "Symmetry and Graph Bi-regularized Non-Negative Matrix Factorization for Precise Community Detection," *IEEE Transactions on Automation Science and Engineering,* DOI: 10.1109/TASE.2023.3240335.

[38] X. Luo, J. P. Sun, Z. D. Wang, S. Li, and M. S. Shang, "Symmetric and Non-negative Latent Factor Models for Undirected, High Dimensional and Sparse Networks in Industrial Applications," *IEEE Transactions on Industrial Informatics,* vol. 13, no. 6, pp. 3098-3107, 2017.

[39] D. Wu, Q. He, X. Luo, M. S. Shang, Y. He, and G. Y. Wang, "A Posterior-neighborhood-regularized Latent Factor Model for Highly Accurate Web Service QoS Prediction," *IEEE Transactions on Services Computing,* vol. 15, no. 2, pp. 793-805, 2022.

[40] D. Wu, and X. Luo, "Robust Latent Factor Analysis for Precise Representation of High-dimensional and Sparse Data," *IEEE/CAA Journal of Automatica Sinica,* vol. 8, no. 4, pp. 796-805, 2021.

[41] X. Y. Shi, Q. He, X. Luo, Y. A. Bai, and M. S. Shang, "Large-scale and Scalable Latent Factor Analysis via Distributed Alternative Stochastic Gradient Descent for Recommender Systems," *IEEE Transactions on Big Data,* vol. 8, no. 2, pp. 420-431, 2022.

[42] D. Wu, P. Zhang, Y. He, and X. Luo, "A Double-Space and Double-Norm Ensembled Latent Factor Model for Highly Accurate Web Service QoS Prediction," *IEEE Transactions on Services Computing,* DOI: 10.1109/TSC.2022.3178543.

[43] H. Wu, X. Luo, and M. C. Zhou, "Advancing Non-negative Latent Factorization of Tensors with Diversified Regularizations," *IEEE Transactions on Services Computing*, vol. 15, no. 3, pp. 1334-1344, 2022.

[44] D. Wu, X. Luo, M. S. Shang, Y. He, G. Y. Wang, and M. C. Zhou, "A Deep Latent Factor Model for High-Dimensional and Sparse Matrices in Recommender Systems," *IEEE Transactions on System Man Cybernetics: Systems,* vol. 51, no. 7, pp. 4285-4296, 2021.

[45] W. L. Li, X. Luo, H. Q. Yuan, and M. C. Zhou, "A Momentum-accelerated Hessian-vector-based Latent Factor Analysis Model," *IEEE Transactions on Services Computing,* DOI: 10.1109/TSC.2022.3177316.

[46] W. L. Li, Q. He, X. Luo, and Z. D. Wang, "Assimilating Second-Order Information for Building Non-Negative Latent Factor Analysis-Based Recommenders," *IEEE Transactions on System Man Cybernetics: Systems,* vol. 52, no.1, pp. 485-497, 2021.

[47] J. Z. Fang, Z. D. Wang, W. B. Liu, S. Lauria, N. Y. Zeng, C. Prieto, F. Sikstrom, and X. H. Liu, "A New Particle Swarm Optimization Algorithm for Outlier Detection: Industrial Data Clustering in Wire Arc Additive Manufacturing," *IEEE Transactions on Automation Science and Engineering,* DOI:10.1109/TASE.2022.3230080.

[48] L Jin, L Wei, S Li, "Gradient-based differential neural-solution to time-dependent nonlinear optimization," *IEEE Transactions on Automatic Control,* vol. 68, no. 1, pp. 620-627, 2023.

[49] X. Luo, W Qin, A. Dong, K. Sedraoui, and M. C. Zhou, "Efficient and High-quality Recommendations via Momentum-incorporated Parallel Stochastic Gradient Descent-based Learning," *IEEE/CAA Journal of Automatica Sinica,* vol. 8, no. 3, pp. 402-411, 2021.

[50] M. Liu, L. Chen, X. Du, L. Jin and M. Shang, "Activated Gradients for Deep Neural Networks," *IEEE Transactions on Neural Networks and Learning Systems,* doi: 10.1109/TNNLS.2021.3106044.

[51] Y. Y. Yao, "The superiority of three-way decision in probabilistic rough set models," *Information Sciences,* vol. 181, pp. 1080-1096, 2011.

[52] Y. Y. Yao, "Three-way decision with probabilistic rough sets," *Information Sciences,* vol. 180, no. 3, pp. 341–353, 2010.

[53] A. Rosenwald, G. Wright, W. C. Chan, J. M. Connors, E. Campo, R. I. Fisher, R. D. Gascoyne, H. K. Muller-Hermelink, E. B. Smeland, and J. M. Giltnane, "The use of molecular profiling to predict survival after chemotherapy for diffuse large-B-cell lymphoma," *New England Journal of Medicine,* vol. 346, no. 25, pp. 1937-1947, 2002.

[54] O. Chapelle, B. Scholkopf, and A. Zien, "Semi-supervised learning (Chapelle, O. *et al*., Eds.; 2006)[Book reviews]," *IEEE Transactions on Neural Networks,* vol. 20, no. 3, pp. 542-542, 2009.

[55] P. Zhou, P. P. Li, S. Zhao, and X. D. Wu, "Feature Interaction for Streaming Feature Selection," *IEEE Transactions on Neural Networks and Learning Systems,* vol. 32, no. 10, pp. 4691-4702, 2021.

[56] P. Zhou, S. Zhao, Y. T. Yan and X. D. Wu, "Online Scalable Streaming Feature Selection via Dynamic Decision," *Association for Computing Machinery,* vol. 16, no. 5, pp. 1556-4681, 2022.

[57] K. Yu, W. Ding, and X. D. Wu, "LOFS: Library of online streaming feature selection," *Knowledge-Based Systems,* vol. 113, pp. 1–3, 2016.

[58] D. Wu, B. Sun, and M. Shang, Hyperparameter Learning for Deep Learning-based Recommender Systems, *IEEE Transactions on Services Computing*, 2023. doi: 10.1109/TSC.2023.3234623.